\documentclass[10pt,twocolumn,letterpaper]{article}

\usepackage{utils/cvpr}
\usepackage[dvipsnames]{xcolor}

\usepackage{array}

\usepackage{multirow}
\usepackage{pifont}
\usepackage{minitoc}
\def \GlobalTableRescale {.98}
\definecolor{darkergreen}{RGB}{21, 152, 56}
\definecolor{red2}{RGB}{252, 54, 65}

\definecolor{maroon}{cmyk}{0,0.87,0.68,0.32}
\definecolor{aliceblue}{rgb}{0.94, 0.97, 1.0}

\definecolor{citecolor}{HTML}{0071BC}
\definecolor{linkcolor}{HTML}{ED1C24}

\definecolor{LightCyan}{rgb}{0.92,1,1}
\definecolor{Gray}{gray}{0.9}

\definecolor{lightRed}{rgb}{0.8,0,0}
\definecolor{lightGreen}{rgb}{0,0.8,0}

\definecolor{deemph}{gray}{0.6}

\newcolumntype{x}[1]{>{\centering\arraybackslash}p{#1pt}}
\newcolumntype{y}[1]{>{\raggedright\arraybackslash}p{#1pt}}
\newcolumntype{z}[1]{>{\raggedleft\arraybackslash}p{#1pt}}

\newlength\savewidth

\newcommand{\ModelName}{CosMo}
\newcommand{\ModelNameTwoB}{CosMo-2B}
\newcommand{\ModelNameThreeB}{CosMo-3.4B}
\newcommand{\ModelNameEightB}{CosMo-8.1B}

\newcommand{\VideoDatasetName} {Howto-Interlink7M}

\definecolor{LightCyan}{rgb}{0.92,1,1}
\definecolor{navy}{RGB}{0,0,128} 
\definecolor{darkgreen}{RGB}{0,100,0}
\definecolor{firebrick}{RGB}{178,34,34}
\definecolor{indigo}{RGB}{75,0,130}
\definecolor{sienna}{RGB}{160,82,45}
\definecolor{darkcyan}{RGB}{0,139,139}
\definecolor{maroon}{RGB}{128,0,0}
\definecolor{darkorchid}{RGB}{153,50,204}
\definecolor{saddlebrown}{RGB}{139,69,19}
\definecolor{darkslategray}{RGB}{47,79,79}
\usepackage{colortbl}

\definecolor{titlecolor}{rgb}{0.6,0.4,0.8}
\newcommand\cosmocolor[1]{{\textcolor{titlecolor}{#1}}}
\definecolor{softblue}{rgb}{0.8, 0.83, 1}
\usepackage{nicematrix} %
\definecolor{background}{rgb}{1,1,1} 

\definecolor{cvprblue}{rgb}{0.21,0.49,0.74}
\usepackage[pagebackref,breaklinks,colorlinks,citecolor=cvprblue]{hyperref}
\title{
    \begin{minipage}{\linewidth}
        \centering
        \begin{minipage}[c]{0.08\linewidth} 
            \includegraphics[width=\linewidth]{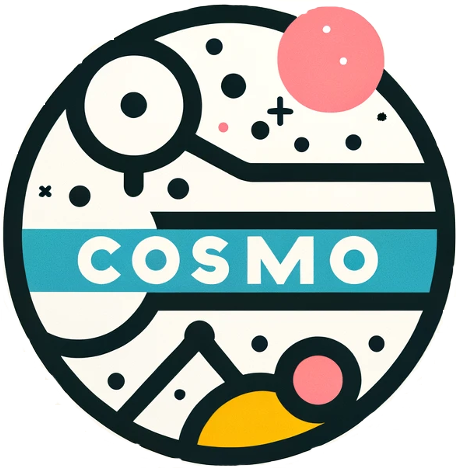} 
        \end{minipage}
        \begin{minipage}[c]{0.85\linewidth} 
            \centering
            \cosmocolor{$\mathcal{COSMO}$}: \cosmocolor{$\mathcal{C}$}\cosmocolor{$\mathcal{O}$}ntrastive \cosmocolor{$\mathcal{S}$}treamlined \cosmocolor{$\mathcal{M}$}ultim\cosmocolor{$\mathcal{O}$}dal Model with \\Interleaved Pre-Training
        \end{minipage}
    \end{minipage}
}

\author{Alex Jinpeng Wang$^{1}$ \quad Linjie Li$^{2}\;$ \quad Kevin Qinghong Lin$^{1}$\quad Jianfeng Wang$^{2}$ \\
Kevin Lin$^{2}$ \quad Zhengyuan Yang$^{2}$ \quad Lijuan Wang$^{2}$ \quad Mike Zheng Shou
$^{1}$\\[3pt]
$^1$Show Lab, National University of Singapore \quad $^2$Microsoft Azure AI \quad \\
\url{http://fingerrec.github.io/cosmo}}

\begin{document}
\maketitle
\begin{abstract}
In the evolution of Vision-Language Pre-training, shifting from short-text comprehension to encompassing extended textual contexts is pivotal. 
Recent autoregressive vision-language models like \cite{flamingo, palme}, leveraging the long-context capability of Large Language Models, have excelled in few-shot text generation tasks but face challenges in alignment tasks.
Addressing this gap, we introduce the contrastive loss into text generation models, presenting the COntrastive-Streamlined MultimOdal framework (\ModelName), strategically partitioning the language model into dedicated unimodal text processing and adept multimodal data handling components. 
\ModelName, our unified framework, merges unimodal and multimodal elements, enhancing model performance for tasks involving textual and visual data while notably reducing learnable parameters. 
However, these models demand extensive long-text datasets, yet the availability of high-quality long-text video datasets remains limited. 
To bridge this gap, this work introduces \VideoDatasetName, an inaugural interleaved video-text dataset featuring comprehensive captions, marking a significant step forward. 
Demonstrating its impact, we illustrate how \VideoDatasetName{} enhances model performance in image-text tasks. 
With 34\% learnable parameters and utilizing 72\% of the available data, our model demonstrates significant superiority over OpenFlamingo~\cite{openflamingo}.
For instance, in the 4-shot flickr captioning task, performance notably improves from 57.2\% to 65.1\%.
The contributions of \ModelName{} and \VideoDatasetName{} are underscored by notable performance gains across 14 diverse downstream datasets encompassing both image-text and video-text tasks.
\end{abstract}    
\begin{figure}
    \centering
    \includegraphics[width=\linewidth]{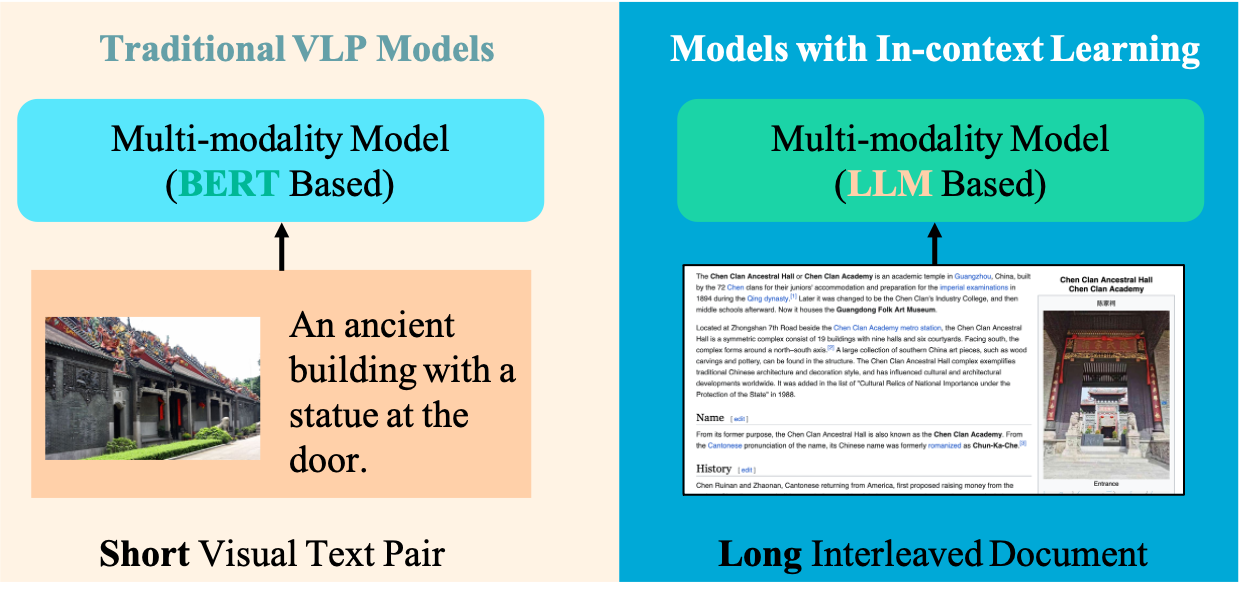}
    \caption{
\textbf{Advancements in Vision-Language Pre-training (VLP) have transitioned towards accommodating long-form text inputs}.
(a). Earlier studies emphasized short, paired image/video text correlations, exemplified by works such as CLIP~\cite{clip} and GiT~\cite{git}.
(b). Present research emphasizes in-context learning strategies, showcased by approaches like Flamingo~\cite{flamingo} and Palm-E~\cite{palme}.
LLMs' exceptional text-processing enables effortless integration of lengthy documents, showcasing robust few-shot learning sans extensive fine-tuning.
    }
    \label{fig:motivation}
\end{figure}

\section{Introduction}
\label{sec:intro}


The emergence of Large Language Models (LLMs)~\cite{opt,redpajama,gpt4} has significantly propelled the development of multi-modal learning paradigms.  
A notable advantage lies in the capacity of LLMs to effectively process extensively lengthy textual inputs, with strong reasoning capabilities~\cite{memorizingtransformer,scalingup}. 
This ability represents a significant stride forward in the domain of Natural Language Processing, underscoring the potential of LLMs in addressing complex, multi-dimensional data. 
The success of LLMs has spurred considerable interests and efforts in leveraging it for multi modalities. 



In-context learning~\cite{gpt3,iclsurvey} provides a possible pathway for models to accept long text inputs
in the realm of multi-modal learning.
Recent advancements in employing in-context learning within multi-modal LLMs have catalyzed the development of Unified Models with Emerging Capabilities, exemplified by Flamingo~\cite{flamingo} and PALM-E~\cite{palme}, showcased in Figure~\ref{fig:motivation}.
These unified frameworks offer the remarkable ability to address numerous downstream tasks without fine-tuning. 
This capability is partly attributed to their architectural design, which supports the utilization of multiple image-text pairs as input and organizes the data into an ``interleaved'' format. 
While these models have exhibited remarkable success in tasks such as Visual Question Answering (VQA) and Captioning, the architecture proposed by Flamingo~\cite{flamingo} is not optimally suited for classification tasks 
as its inherent design focuses on text generation rather than classification. 

High-quality interleaved data is required to enable models with multimodal in-context learning.
However, the majority of publicly available datasets, such as CC3M~\cite{cc3m}, LAION400M~\cite{laion400m}, and DataComp1B~\cite{datacomp}, predominantly consist of short image-text pairs.
Recent efforts by CM3 dataset~\cite{cm3}, MMC4~\cite{mmc4} and Obelics~\cite{obelics} have introduced three publicly accessible interleaved datasets, based on web documents.  
However, web documents can be noisy, as the images on the same page might not be highly correlated (see Figure~\ref{fig:lowsimsample}).
Compared to web documents, videos, naturally encompass highly-correlated image sequences.
In response, initiatives like InternVid~\cite{internvid} and Video Chapters~\cite{vidchapters} have proposed longer captions generated by language models or user-annotated chapters.
Despite these advancements, the availability of interleaved video-text data, demonstrating relationships across different clips, remains deficient.

To address these challenges, this paper introduces a novel architecture capable of processing four distinct types of inputs, including interleaved data, aiming to rectify the limitations observed in Flamingo. 
Our approach involves dividing the LLM into two segments: the first segment specializes as a text encoder, while the second part is used for multimodal fusion.
Additionally, 
we present \VideoDatasetName, a high-quality interleaved video-text dataset derived from Howto100M~\cite{howto100m}, by leveraging GPT-4~\cite{gpt4}. 
\ModelName{} is evaluated across a total of 14 image-text and video-text benchmarks, achieving superior performance compared to Open-Flamingo~\cite{openflamingo}, while utilizing fewer samples from the same public datasets. Our results further show that high-quality video-text data from our \VideoDatasetName{} further enhances the performance of \ModelName{}, even helps on image-text tasks.

Our key contributions include:
\emph{(i)}. We introduce a novel architecture \ModelName{} for interleaved data pre-training, leveraging an additional contrastive loss.
With only 34\% learnable parameters, our method outperform~\cite{openflamingo} clearly.
\emph{(ii)}. We introduce \VideoDatasetName{}, a noteworthy addition to long text multi-modality datasets.
\emph{(iii)}. We show that top-tier interleaved video-text data boosts model performance in various image-text and video-text tasks.

\section{Related Work}
\label{sec:formatting}

\paragraph{Vision-Language Pretraining.} The evolution of Language Modeling has significantly impacted vision-language pre-training methodologies. 
Traditional approaches such as OSCAR~\cite{oscar}, ViLT~\cite{vilt} and UNITER~\cite{uniter}, built upon BERT~\cite{bert} language architectures, have demonstrated prowess in downstream tasks without requiring extensive fine-tuning.
The focus, however, pivots towards harnessing larger language models~\cite{forzen,blip2}. 
For instance, the Flamingo~\cite{flamingo} has escalated from a 1.4B to a staggering 70B parameter language model, showcasing robust performance across various downstream tasks. 
Notably, Flamingo's architecture is adept at handling interleaved image/video text datasets through specialized input formats and cross-attention mechanisms. 
Yet, its performance falls behind contrastive models in classification tasks, a limitation compared with other works like CLIP~\cite{clip} and CoCa~\cite{coca}, which thrive on contrastive learning paradigms.

To capitalize on these insights, we adopt Flamingo's input design and incorporate the contrastive loss within middle-level LLM representations, to further enhance the alignment between visual and text representations. 


\paragraph{Interleaved Data for Multi-modality Learning.}
Acquiring manually annotated vision-language datasets is prohibitively expensive, leading to relatively small-scale datasets (often smaller than 100k instances) such as COCO~\cite{coco} and Visual Genome~\cite{vg}. 
Traditionally, vision-language datasets are mainly composed of image-text pairs from Internet, such as CC3M~\cite{cc3m} and WIT in CLIP~\cite{clip}.
The text captions in these web-crawled datasets, are mostly alt-text descriptions, which are mostly short in length, hence less descriptive.
An innovative shift was introduced by Flamingo~\cite{flamingo}, pioneering the concept of long-form image-text pairs. 
Recent advancements from Flamingo~\cite{flamingo} and CM3~\cite{cm3} emphasized the significance of training on entire multimodal webpages, presenting interleaved images and text as a cohesive sequence. These interleaved datasets inherently encompass multiple image and text pairs, contributing to the evolving landscape of multimodal learning. 
The MMC4 dataset~\cite{mmc4} notably stands as the initial publicly available work directly addressing multi-image/multi-sentence interleaved data. 
However, MMC4 is missing a crucial detail, the exact placements of images within the document structure, which is addressed in obelics~\cite{obelics}.

Existing research has predominantly concentrated on extracting highly correlated image-text data from noisy web documents. 
In contrast, our work introduces a pioneering interleaved video-text dataset, marking a departure from solely image-text modalities.

\begin{table}[]
    \centering
    \footnotesize
    \begin{tabular}{c|cccccc}
    \toprule
        Datasets & Videos & Clips & Docs  & Token & Sim. \\
        \midrule HowTo100M~\cite{howto100m} & 1.22M & 136.6M  & - & 30 & 0.22 \\
         \VideoDatasetName&1M& 7M & 1M & 55 & 0.32\\ 
    \bottomrule
    \end{tabular}
    \vspace{-.5em}
    \caption{\textbf{Data statistics of \VideoDatasetName}.
    The last two columns are average token length and CLIP~\cite{clip} image-text similarity score for each clip, respectively.
    }
    \vspace{-1em}
    \label{tab:vlog_details}
\end{table}

\begin{figure*}
    \centering
\includegraphics[width=.85\linewidth]{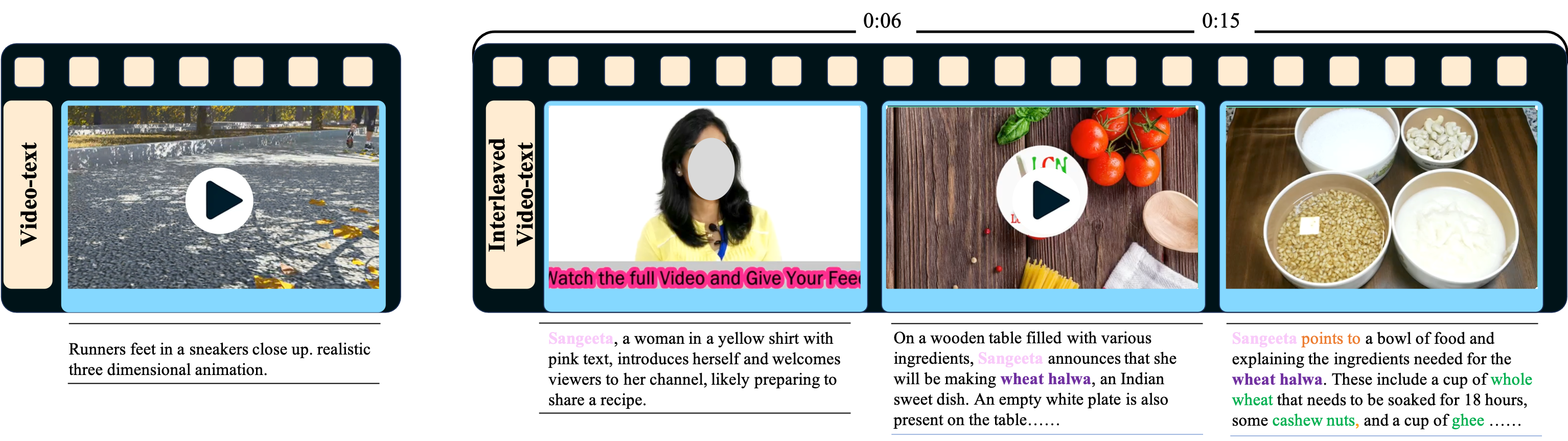}
    \vspace{-1em}
    \caption{
    \textbf{Comparing Conventional Video-Text Datasets with Our \VideoDatasetName.}
    Left: Conventional video-text datasets typically contain brief ASR captions describing videos.
    Right: In \VideoDatasetName, to include more details and improved video coherence, videos are segmented into shots.
    Following this, the GPT-4 model~\cite{gpt4} is employed to annotate each shot based on \textbf{historical context} and \textbf{detailed annotations} include ASR, captions, and dense captions.
    We highlight hard object labels as well as connectives between clips.
    }
    \label{fig:vlog_example}
\end{figure*}

\section{\VideoDatasetName}
\paragraph{Motivation.}
The effectiveness of video-language pre-training often hinges on the availability of high-quality annotated captions. 
Existing datasets such as Howto100M~\cite{howto100m} and YT-Temporal~\cite{ytt} predominantly rely on YouTube videos with texts generated by Automatic Speech Recognition (ASR). However, these ASR-generated annotations suffer from weak alignments with the video content. As observed in Howto100M,  only 51\% of clips feature objects or actions mentioned in their captions are visibly depicted in the sampled clips~\cite{howto100m}. 
This limitation poses a significant challenge to the vision-language pre-training community.
To address this issue, we introduce \VideoDatasetName, a novel interleaved video-text dataset aiming to provide high-quality annotations for improved video-language understanding.

\paragraph{Generation Pipeline.}
To construct \VideoDatasetName, we start with the publicly available HowTo100M dataset, but undertake a distinctive approach. 
Diverging from the original dataset, we segment the original videos into shots using a shot detector, specifically KTS~\cite{kts}. 
For each shot, we employ off-the-shelf caption models like BLIP2~\cite{blip2} to generate concise descriptions. 
Furthermore,  we employ GRIT~\cite{grit} to produce dense region captions with bounding box, enhancing the captions with rich descriptions and position details.
For simplicity, we name all these information as \textbf{detailed annotation}. 
For first clip of video, we leverage GPT-4~\cite{gpt4} model to generate comprehensive summaries from \textbf{detailed annotation} only.
Importantly, subsequent clips' captions are conditioned on the context of preceding clips, maintaining narrative continuity. 
We provide explicit instructions to GPT-4, emphasizing the preservation of ASR information to retain the nouns and actions.
This approach strengthens the interconnection between individual clips, fostering a more coherent caption. 

\paragraph{Analysis.}
In Figure~\ref{fig:vlog_example}, we showcase an example illustrating the comprehensive and detailed nature of the generated paragraphs. 
Notably, our observations indicate the preservation of specific details across different clips, including the names of actors and mentioned objects like food.
Moreover, we conduct a comparative analysis between the original HowTo100M~\cite{howto100m} dataset and our \VideoDatasetName, as presented in Table~\ref{tab:vlog_details}. 
The generated captions exhibit greater length and improved alignment with the video content, as evidenced by the higher CLIP similarity scores. 

\vspace{-.8em}

\begin{table}[]
    \centering
    \footnotesize
\scalebox{.95}{
    \begin{tabular}{l|lll}
    \toprule
        \bf Method & \bf Language model & \bf Vision Model & \bf \#CE  \\
        \midrule
        \ModelNameTwoB & OPT-IML1.8B~\cite{optiml} & Open-Clip ViT-L/14~\cite{openclip} & 6  \\
        \ModelNameThreeB & RedPajama-3B~\cite{redpajama} & Open-Clip ViT-L/14~\cite{openclip} & 4\\
        \ModelNameEightB & Mistral7B~\cite{mistral} & Open-Clip ViT-L/14~\cite{openclip} & 4 \\
        \bottomrule
    \end{tabular}
    }
    \vspace{-.5em}
    \caption{\textbf{Architecture details of the \ModelName}.
    \#CE is short for the number of Cross Attention layers.}
    \vspace{-2em}
    \label{tab:my_label}
\end{table}

\section{\ModelName}
 In this section, we introduce \ModelName, short for COntrastive-Streamlined MultimOdal framework, which strategically partitioning a LLM into dedicated unimodal text processing and adept multimodal data handling components.
 \ModelName{} adds an additional contrastive loss to the language model loss in Flamingo~\cite{flamingo,openflamingo} baselines, supporting both classification and generation tasks.
While ALBEF~\cite{albef} and CoCa~\cite{coca} also integrate contrastive loss, our emphasis lies in developing LLMs specialized in in-context learning and handling extensive interleaved data sequences, setting our work apart from these methods.
Additionally, we streamline our model architecture, reducing the number of learnable parameters for improved computational efficiency while maintaining efficacy in multi-modality learning.

\subsection{Overall Architecture}
\begin{figure}[t]
  \centering
\includegraphics[width=\linewidth]{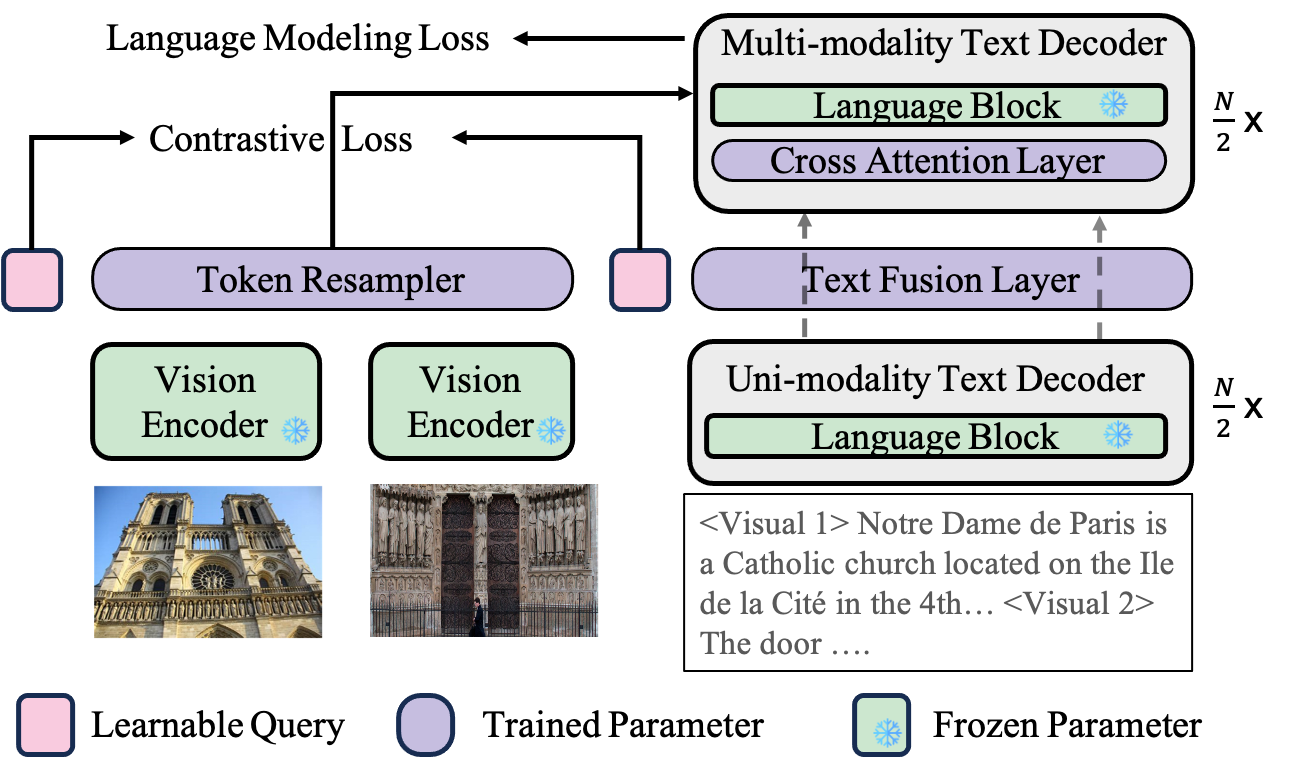}
   \caption{
   \textbf{An introduction to \ModelName}: This model handles both image/video text pairs and inter-level image/video text pairs. 
    The Large Language Model is divided into two parts to compute contrastive loss and language modeling loss.
   }
   \label{fig:main_ppl}
\end{figure}

As shown in Figure~\ref{fig:main_ppl}, \ModelName{} consists of two components: a visual encoder and a pre-trained LLM. 
The visual encoder, based on the Vision Transformer (ViT)~\cite{vit} from Open-CLIP~\cite{openclip}, remains consistent across our experiments. For the language model, we primarily employ the OPT~\cite{opt} model, partitioned into two segments to accommodate diverse tasks while reducing overall parameters.

\paragraph{Input Representation as Document.}
\ModelName{} accepts four types of data: image-text, video-text, interleaved image-text, and interleaved video-text, all processed into a document-style text sequence format. 
This format encapsulates visual and textual information, structured as ``\textless s\textgreater\textless Visual 1\textgreater Text1 \textless EOC\textgreater \textless Visual 2\textgreater Text2 \textless EOC\textgreater'', with \textless s\textgreater~marking the document start and \textless EOC\textgreater~denoting the end of each text caption. 
The \textless Visual\textgreater~token represents the presence of an image or video.

To effectively handle longer documents while constrained with a fixed GPU memory budget, we implement a random sampling strategy to gather inputs with a fixed number of tokens.
We first determine the position of the \textless Visual \textgreater~token, and randomly select one image as the anchor. Subsequently, we introduce a slight random shift to the image position, and then sample 128 tokens subsequently. The images or videos corresponding to the \textless Visual \textgreater~tokens serves as inputs to the visual encoder.

\paragraph{Uni-modality Feature Extraction.}
To mitigate catastrophic forgetting, inspired by recent works~\cite{magma,flamingo}, we freeze both the LLMs and the vision encoder. 
Images or videos are directly fed into the frozen vision encoder, while documents are input into the first half layers of the LLM.

\paragraph{Lightweight Multi-modal Fusion.}
Then, we leverage visual tokens to condition the frozen language model block through gated cross-attention layers, which share a similar design with Flamingo~\cite{flamingo} and CoCa~\cite{coca}. 
This strategy effectively integrates visual information for precise next-token prediction. 
However, a key difference from previous methods is that we introduce bottlenecks~\cite{inception,resnet} in input and output feature channels, resulting in a substantial reduction in learnable parameters.
Specifically, we compress the feature channel dimension to one half at the beginning and raise it again at last.
Moreover, cross-attention layers are strategically introduced at regular intervals, specifically every 2 blocks for \ModelNameTwoB, and 4 for \ModelNameThreeB.

\begin{figure}
    \centering
    \includegraphics[width=\linewidth]{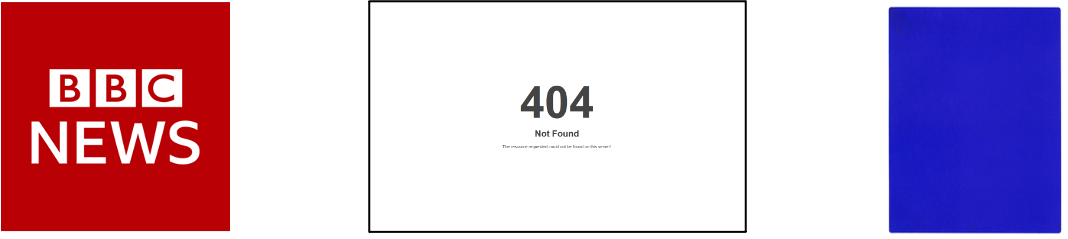}
    \caption{
    \textbf{Instances of Low-Similarity Images}:
    In datasets like MMC4~\cite{mmc4}, based on raw website content, there are often incongruent images that don't align with the accompanying text, leading to training instability.
    }
    \label{fig:lowsimsample}
\end{figure}

\paragraph{Loss:}For document feature $x$ and vision feature $z$, the model predicts the next word $y$ and the loss is computed as:
\begin{equation}
p(y|x) = -\sum_{i=1}^{i=T} \log p(y|x,z),
\end{equation}
where $T$ is the length of the input text. The final loss function is composed of both language modeling loss ($\mathbf{L}_l$) and contrastive loss ($\mathbf{L}_c$):
\begin{equation}
\mathbf{L} = \sum_{i=1}^{i=N} \lambda_1 \mathbf{L}_l + \lambda_2 \mathbf{L}_c,
\end{equation} where $N$ is the number of data types (3 by default).

To this end, our \ModelName{} is trained to adeptly handle interleaved text and visual sequences, naturally enabling its application in in-context few-shot learning scenarios.

\subsection{Enhancing Image-Text Alignment}
The Vision Encoder generally stems from the CLIP~\cite{clip} model, meticulously trained with contrastive loss to preserve rich and discriminative information.
Preceiver resampler~\cite{flamingo} obtains spatio-temporal features from the Vision Encoder, yielding a fixed number of visual tokens. 
But the risk of missing discriminative details persists, potentially leaving certain information unclear.

Contrarily, our approach extends this process by incorporating a learnable query to globally attend to all tokens, including an additional learnable query for Text Fusion Layers. 
This modification allows us to attain a more comprehensive understanding of the entire token set. 
Subsequently, the model employs a projection head to unify visual and text embeddings into the same dimensionality. 
The training objective centers around optimizing the contrastive loss. 

\subsection{Interleaved Data Preprocessing}
\label{sec:interleave_preprocess}
\paragraph{Image-text Similarity Matrix.} In the interleaved MMC4~\cite{mmc4} dataset, each document (typically a website) contains a text list and an image list extracted from the document. 
Also, this dataset provides pairwise image-text similarity computed using the CLIP~\cite{clip} model. 
Due to the absence of image positions within the document, during pre-training, Open-Flamingo~\cite{openflamingo} select the matching index utilizing Optimal Transport~\cite{optimaltransport} from the similarity matrix.

\begin{table}[]
    \centering
    \footnotesize
    \begin{tabular}{l|ll}
    \toprule
        Data Type & Dataset & Sample  \\
        \midrule
        \multirow{4}{*}{Image-Text} & CC3M~\cite{cc3m} & 0.75M \\
        &SBU~\cite{sbu} & 0.3M\\
        &LAION400M~\cite{laion400m} & 30M\\
        &DataComp1B~\cite{datacomp} & 65.6M\\
        \hline
        Interlevel Image-Text&MMC4~\cite{mmc4} &  30M \\
        \hline
        Video-Text & WebVid~\cite{webvid} & 2.5M \\
        \hline
        Interlevel Video-Text & \VideoDatasetName  & 0.9M \\
        \midrule
        Total & - & 130M \\
        \bottomrule
    \end{tabular}
    \vspace{-1em}
    \caption{
    \textbf{Statistics of the Pre-training Dataset}:
    We extract a subset from the complete dataset using clustering and filtering.
    }
    \label{tab:dataset_details}
\end{table}

\paragraph{Data Filtering.} The MMC4 dataset comprises numerous images with low similarity scores, which may include logos, failed image downloads, or images completely unrelated to the accompanying text, as shown in Figure~\ref{fig:lowsimsample}. 
Training models directly on such data often results in gradient explosions due to the inherent irrelevance between images and text. 
To mitigate this issue, MMC4 employs a simple filtering using a threshold of similarity score 0.24, computed by CLIP ViT-L/14.
However, this approach discards a significant number of samples, reducing the dataset's diversity.

To overcome these limitations, we implement the following strategies:
\emph{(i.)} \textbf{Similarity Distribution}: Unlike previous methods, we introduce a disturbance matrix with a normal distribution having a standard deviation within the range $(-0.04, 0.04)$ for the alignment score matrix. 
Furthermore, we clamp the min and max value to -0.08 and 0.08, correspondingly.
We then utilize optimal transport for index matching. This approach allows images to match with different texts within the same document.
\emph{(ii.)} \textbf{Preserving Document Context}: For images with a similarity score below 0.20, we leverage a pre-trained captioning model to generate pseudo-captions to replace the matched texts within the document. This operation significantly diversifies the sampled documents and enhances the training stability. 
Refer to Section~\ref{sec:interleave_sampling_explore} for comprehensive analysis.

\subsection{Training Details}

\textbf{Pre-training Data:}
Conventional datasets often suffer from noise and duplication issues. 
Following the approach outlined in TLDR~\cite{tldr}, we opt to sample a refined subset from commonly used image-text pre-training datasets. 
Table~\ref{tab:dataset_details} details the specific datasets sampled for our model training. 
Further analysis on our data sampling methodology is shown in Section~\ref{sec:setting}.

\paragraph{Training Configuration:}
All models are trained using AdamW~\cite{adamw} optimizer with a cosine learning rate schedule, initialized at 5e-4, and trained for 20 epochs. 
The warm-up ratio is set to 0.03. 
Gradient accumulation steps align with the data type, defaulting to 3 for our \ModelName. 
We employ DeepSpeed~\cite{deepspeed} fp16 precision for model training, executed across 128 NVIDIA V100 GPUs.

\section{Experiments} 

Our primary objective is to develop models capable of swift adaptation to a wide array of diverse and challenging tasks. To this end, we evaluate our model against numerous well-established image-text~\cite{coco,flickr30k,okvqa,textvqa,vizwiz,vqav2} and video-text~\cite{tvc,msvd,msrvtt,youcook2,vatex,tgif} benchmarks, to assess the few-shot in-context learning performance of \ModelName{}.


\begin{figure}[t]
  \centering
    \includegraphics[width=.9\linewidth]{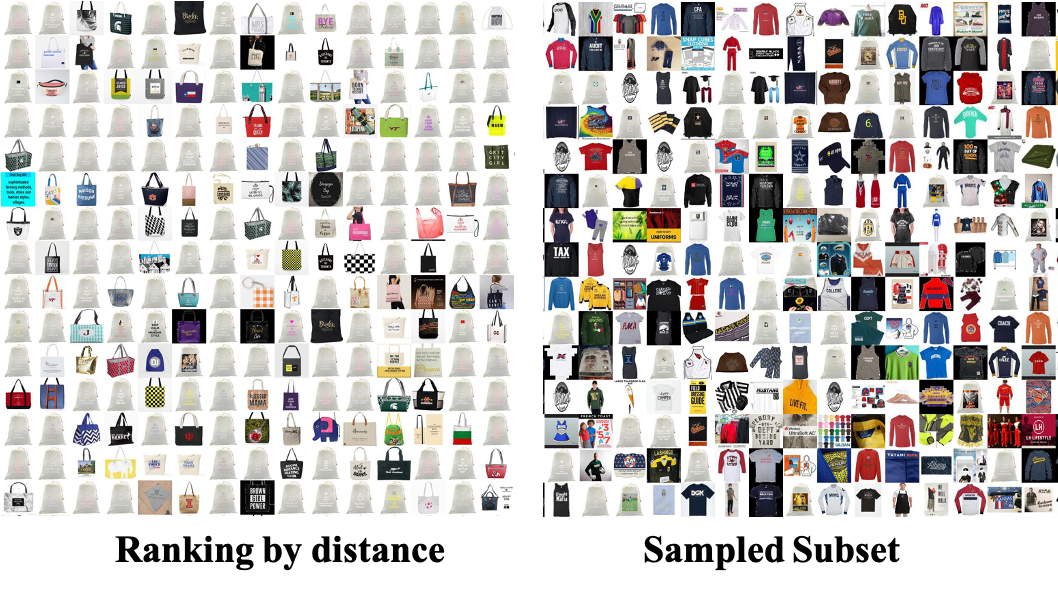}
       \vspace{-1em}
   \caption{
   \textbf{
   LAION400M~\cite{laion400m} and similar large datasets commonly suffer from redundancy.}
    Clustering and uniform distance-based sampling help alleviate this issue.
   }
   \vspace{-1.2em}
   \label{fig:cluster_visualization}
\end{figure}

\subsection{Pre-training Data Selection}
\label{sec:setting}

\begin{table*}[]
    \centering
    \footnotesize
    \scalebox{\GlobalTableRescale}
{
    \begin{tabular}
    {y{55}y{37}y{50}x{37}x{17}|x{17}x{17}|x{24}x{17}x{17}x{17}|x{22}}
    \toprule
    \bf Ablated setting & \bf Original & \bf Changed &\bf Parameter & \bf Iter. & \multicolumn{2}{c}{\bf Captioning (CIDER)}  & \multicolumn{4}{c}{\bf VQA}& \bf Average  \\
    & Value& Value && Time & COCO & FLICKR & ok-vqa & textvqa & vizwiz & vqav2  & \\
    \midrule
    \multicolumn{3}{c}{\bf Baseline} & 683M/2.32G & 4.9s & 71.3 & 50.9 & 18.5 &15.0& 33.6& 38.4& 38.0\\
    \hline
    \bf \multirow{2}{*}{\parbox[c]{2cm}{i. Contrastive Loss}} &  \multirow{2}{*}{wo Contrastive} & \textcolor{indigo}{Single GPU} & 683M/2.32G&5.1s & 76.0 & 53.0 &20.9& 16.7& 35.5& 41.0& 40.5\\
    &&All GPUS & 683M/2.32G & 6.1s & 73.1 & 52.0 & 20.4 &14.0 &26.4 &39.4 & 39.7\\
    \hline
    \bf \multirow{3}{*}{\parbox[c]{1.5cm}{ii. Training Data}} & \multirow{3}{*}{\textcolor{indigo}{All Data}} & w/o Video-text
    & 683M/2.32G&4.6s & 68.8 & 51.6 & 16.2 & 14.1 & 30.7   & 35.3 & 36.1 \\
    &&w/o Interleaved&683M/2.32G & 3.3s & 42.1 & 26.0 &9.5& 6.1& 9.4& 27.2& 20.1\\
    &&w/o Image-text&683M/2.32G & 4.5s &44.1&26.3&9.6& 4.1& 6.3& 32.7 & 18.5\\
    \hline
    \bf \multirow{2}{*}{\parbox[c]{2cm}{iii. Dataloader Sampling} } & \multirow{2}{*}{Round-robin} & \textcolor{indigo}{Min} &683M/2.32G & 4.9s & 73.2 &56.9& 22.1& 15.8& 33.7& 42.0 & 40.6\\
    &&Max&683M/2.32G& 4.9s &73.2 & 53.2 & 19.0 & 16.4& 20.0 &41.0 &37.1\\
    \hline
    \bf \multirow{2}{*}{\parbox[c]{2cm}{\parbox[c]{1.5cm}{iv. Vision Encoder}}} & \multirow{2}{*}{\textcolor{indigo}{ViT-L}} &ViT-B
    & 617M/1.98G & 4.2s & 64.7 & 45.9 &19.3& 13.5& 21.4& 39.0 & 34.0\\
    &&ViT-G & 700M/3.28G & 7.5s & 72.1 & 54.1& 20.6& 19.7 &34.0& 41.5&  40.3\\
    \hline
    \bf \multirow{2}{*}{\parbox[c]{2cm}{v. Layer Inter}} & \multirow{2}{*}{1} &\textcolor{indigo}{2}
    & 462M/2.10G & 4.5s & 71.4 & 53.8 & 22.3& 16.4 &28.5 &42.3 & 39.1 \\
    &&4 &352M/1.99G & 3.9s & 70.4 &51.3& 20.8 &15.8& 30.7 &41.8 &38.5\\
    \hline
    \bf \multirow{2}{*}{\parbox[c]{2cm}{vi. Compress Ratio}} & \multirow{2}{*}{1} &\textcolor{indigo}{2}
    & 444M/2.08G & 4.3s & 71.5 & 54.9 &22.1& 14.7& 31.7& 41.1 & 39.3 \\
    &&4& 325M/1.97G &4.2s&72.0 & 48.0 &20.8& 12.8& 32.2& 41.3 & 37.9 \\
        \hline
    \bf \multirow{2}{*}{\parbox[c]{2cm}{vii. Interleaved Length}} & \multirow{2}{*}{64} &\textcolor{indigo}{128}
    & 683M/2.32G & 6.5s & 73.1 & 53.5 &22.8& 15.7& 33.4& 40.8  & 39.9\\
    &&192 & 683M/2.32G & 9.3s & 72.8 &54.5 &22.5& 14.9& 33.4 &42.0 & 40.0\\
    \bottomrule
    \end{tabular}
    }
    \caption{
    \textbf{Ablation Studies of \ModelName~on an 18M Subset}.
    Our focus is primarily on presenting the 8-shot results for comparison, with the baseline result positioned in the first row.
    We highlight the distinction between learnable and all parameters.
    For the captioning assessment, we report CIDER, and 'Iter.' abbreviates Iteration.
    Text in \textcolor{indigo}{indigo} denotes our default setting.
    }
    \label{tab:ablation}
\end{table*}

Our experimental setup draws inspiration from TLDR~\cite{tldr}, where we leverage a subset of the SBU~\cite{sbu}, CC3M~\cite{cc3m}, CC12M~\cite{cc12m}, LAION400M~\cite{laion400m}, and Data-comp 1B~\cite{datacomp} datasets. To obtain this subset, we begin by filtering out half of the data with low similarity scores. Subsequently, we employ K-means clustering and uniformly sample data from each cluster, resulting in a total of 100 million data points. 
We compute the centroid of each cluster and uniform sample data according to the distance from the mcenter.
An illustrative example of this clustering process is shown in Figure~\ref{fig:cluster_visualization}.
For interleaved data, we utilize 30 million data points from MMC4~\cite{mmc4} following a filtering process by removing samples with too small images or no images. 
Also the data filtering in Section~\ref{sec:interleave_preprocess}.

\subsection{Ablation Study}
\label{sec:ablation_study}

In Table~\ref{tab:ablation}, we report our ablation results using our model pre-trained on a 18M subset (12M Interleaved, 4M Image and 2M Video-text).
The average score is computed by averaging the scores across all datasets.

\paragraph{Significance of Contrastive Loss:}
As seen in section (i) of Table~\ref{tab:ablation}, the contrastive loss plays a critical role in the success of our approach. 
We observe that using the contrastive loss on both single and multiple GPUs consistently improves model performance over the baseline.
Yet, our further investigations indicate that aggregating all contrastive losses across nodes does not provide commensurate benefits. This can be attributed to the model's strong focus on the contrastive loss, which can affect the learning of language modeling loss.
In addition, employing the contrastive loss on all GPUs introduces a significant lag in training speed (from 4.9s to 6.1s), which increases computation expenses. 
As a result, by default we use the contrastive loss over batch on each GPU.

\begin{table*}
\centering
\footnotesize
\begin{NiceTabularX}{\linewidth}{y{52}x{20}x{20}x{20}x{17}|cc|cccc|c|c}
\CodeBefore
\rowlistcolors{1}{background,background,background,background,background,background,background,background,softblue,softblue,softblue,softblue,softblue,softblue,background,background,background,softblue,softblue,softblue,background,background,background,background,background,background,softblue,softblue,softblue}
\Body
  \footnotesize
    \bf Method  & \bf Samples & \bf Para. & \bf \#Tokens$\downarrow$&\bf Shots & \multicolumn{2}{c}{\bf Captioning (CIDER)}  & \multicolumn{4}{c}{\bf VQA} 
    &\bf Classification& \bf Avg $\uparrow$ \\
    & & & & & COCO & FLICKR & ok-vqa & textvqa & vizwiz & vqav2 & hatefulmemes & \\
    \midrule
    \bf \multirow{3}{*}{\parbox[c]{2cm}{Flamingo-3B~\cite{flamingo}}} &  \multirow{3}{*}{2.1B} & \multirow{3}{*}{1.4B/3.2B} & \multirow{3}{*}{256}
    & 0& 73.0 & 60.6& 41.2 & 30.1 & 28.9 &  49.2 & 53.7& 48.1 \\
    &&&& 4 & 85.0 & 72.0 & 43.3   & 32.7 & 34.0 &53.2 & 53.6 & 53.4\\
    &&&& 32 &  99.0 & 71.2 & 45.9 & 30.6 & 45.5 & 57.1 &56.3 & 57.9\\
    \hline
    \bf \multirow{3}{*}{\parbox[c]{2cm}{Open-Flamingo(3B)~\cite{openflamingo}}}   & \multirow{3}{*}{180M} & \multirow{3}{*}{1B/3B} & \multirow{3}{*}{256}
    & 0& 74.9 & 52.3& 28.2 & 24.2 & 23.7 & 44.6 & 51.2 & 42.7  \\
    &&&& 4 & 77.3 & 57.2 & 30.3 & 27.0 & 27.0 & 45.8& 50.6 &  45.0 \\
    &&&& 32 &  93.0 & 61.1 & 31.0 & 28.3 & 39.8 & 47.0 & 50.2 & 50.0 \\
    \hline
    \bf \multirow{3}{*}{\ModelNameTwoB} & \multirow{3}{*}{130M} & \multirow{3}{*}{340M/1.9B} & \multirow{3}{*}{128}
    & 0 & 79.9& 51.3& 28.3 & 21.9 & 24.7 & 46.7 &51.2& 43.4 \\
    &&&&4&91.7&61.2&27.3&23.4&30.0&45.5 &50.6& 47.1 \\
    &&&&32&95.4&64.7 & 26.8 & 25.6 & 42.3 & 43.9 &50.9& 49.9 \\
    \hline
     \bf \multirow{3}{*}{\parbox[c]{2.3cm}{\ModelNameTwoB+\VideoDatasetName }} & \multirow{3}{*}{131M} & \multirow{3}{*}{340M/1.9B} & \multirow{3}{*}{128}
    & 0 & 81.3& 52.8& 25.5 & 27.7 & 27.4 & 45.3 &50.5&44.4  \\
    &&&&4&97.9&65.1&28.7&25.8&30.3& 47.8 &51.8& 49.6 \\
    &&&&32&100.1&63.8 & 25.8 & 25.9 & 42.1 & 44.6 &49.8& 50.3\\
    \midrule
  \bf \multirow{3}{*}{\parbox[c]{2cm}{Open-Flamingo(4B)~\cite{openflamingo}}}   & \multirow{3}{*}{180M} & \multirow{3}{*}{1.3B/4B} & \multirow{3}{*}{256}
    & 0& 76.7 & 53.6& 30.7 & 21.8 & 18.8 & 45.1 & 52.3& 42.7  \\
    &&&& 4 & 81.8 & 60.7 & 35.1 & 25.9 & 26.6 & 49.0 & 51.5 & 47.2  \\
    &&&& 32 &  95.1 & 56.9 & 26.4 & 14.1 & 23.1 & 43.0 & 52.2 & 44.4\\
    \hline
        \bf \multirow{3}{*}{\ModelNameThreeB} & \multirow{3}{*}{130M} & \multirow{3}{*}{405M/2.9B} & \multirow{3}{*}{128}
    & 0 & 78.7 & 55.4 & 32.5 & 23.4 &20.9 & 46.2 & 50.6 & 44.0\\
    &&&&4 & 91.4 & 63.4 & 35.3 & 21.0 & 28.4 & 48.8 & 52.7 & 48.7\\
    &&&&32 &101.4&67.4&30.4&20.8& 35.4& 47.8 & 51.3 & 50.6\\
    \midrule
    \bf \multirow{3}{*}{\parbox[c]{2cm}{Flamingo(9B)~\cite{flamingo}}} & \multirow{3}{*}{2.1B} & \multirow{3}{*}{1.8B/9.3B} & \multirow{3}{*}{256}
    & 0& 79.4&61.5&44.7&31.8&28.8&51.8 & 57.0 & 50.7 \\
    &&&&4 &93.1&72.6&49.3&33.6&34.9&56.3 &62.7 & 57.5\\
    &&&&32 &106.3&72.8&51.0&32.8&44.0& 60.4& 63.5 & 61.5 \\
    \hline
        \bf \multirow{3}{*}{\parbox[c]{2cm}{Open-flamingo(9B)~\cite{openflamingo}}} & \multirow{3}{*}{180M} & \multirow{3}{*}{1.9B/9B} & \multirow{3}{*}{256}
    & 0& 74.9 & 52.3& 28.2 & 24.2 & 23.7 & 44.6 & 51.6 & 42.8  \\
    &&&& 4 & 77.3 & 57.2 & 30.3 & 27.0 & 27.0 & 45.8 & 54.0 &  45.5 \\
    &&&& 32 & 93.0 & 61.1 &  31.0 & 28.3 & 39.8 & 47.0 & 50.2 & 50.1 \\
    \hline
        \bf \multirow{3}{*}{\parbox[c]{2cm}{\ModelNameEightB}}  & \multirow{3}{*}{180M} & \multirow{3}{*}{514M/8.1B} & \multirow{3}{*}{128}
    & 0& 77.5 &58.2&32.7&23.5&22.5&47.2& 57.1 & 45.5 \\
    &&&&4 &85.5&63.4&33.7&26.8&30.5&47.7&58.9 & 50.0\\
    &&&&32 &103.5&67.7&34.0&28.9&41.3& 49.2&62.0 & 55.2 \\
    \bottomrule
\end{NiceTabularX}
    \caption{
    \textbf{Comparison Across Various Scales}.
    By employing only 128 tokens in pre-training, our computational costs are significantly lower compared to related approaches.
    'ILLength' abbreviates Interlevel Token Length.
    It's noteworthy that \ModelNameThreeB~and Open-Flamingo (4B) utilize the same LLM (RedPajama-3B) and visual encoder (Clip ViT-L/14).
    Rows with color indicate our own method.
    }
    \label{tab:sota}
\end{table*}

\paragraph{All Data are important:}
As demonstrated in section (ii) of Table~\ref{tab:ablation}, the selection of appropriate training data is important to the model performance. 
Notably, the removal of the interleaved image-text dataset results in huge performance degradation in terms of the average score (from 38.0 to 20.1), and omitting the conventional image-text pairs similarly affect the performance. Additionally, the exclusion of paired video-text datasets has a detrimental impact on a majority of downstream tasks.

In light of these findings, we employ gradient accumulation, and the number of accumulation steps is set to match the number of data types. 
This ensures that each iteration encompasses all data types, allowing us to leverage the full spectrum of data available for training.
We also present various dataloader sampling strategies in row (iii) of Table~\ref{tab:ablation}. Our findings indicate that the  ``minimum'' strategy outperforms both ``maximum'' and ``round-robin.'' This underscores the importance to ensure balanced quantities across each type of training data. 

\paragraph{Visual Encoder Size:}
In section (iv) of Table~\ref{tab:ablation}, we evaluate the effects of altering the size of the visual encoder. 
Our observations reveals a general trend that 
larger visual encoders tend to produce slightly better results. 
However, this improvement is offset by a corresponding increase in the number of 
model parameters and computational demands. 
Notably, the time required per iteration increases from 1.2 seconds to 2 seconds when using larger visual encoders.
Considering this trade-off between performance gains and computational costs, we have 
chosen to adopt the larger visual encoder size as the default configuration.

\paragraph{Lightweighting the Model:}
In our pursuit of a more lightweight and efficient network, we focus on reducing the learnable parameters by minimizing the number of cross-attention layers and compressing its associated parameters.

The results are presented in sections (v) and (vi). 
Surprisingly, reducing the number of learnable parameters does not generally lead to a performance decrease. 
It even results in enhanced performance, especially when employing a layer interval of 2 and a compression ratio of 2, which we incorporate into our final framework.

\paragraph{Interleaved Length Ablation:}
In section (vii) of Table~\ref{tab:ablation},  we explore the impact of varying the interleaved sequence length within the range of 64 to 192. In general, longer sequences lead to improved results. 
However, it's worth noting that longer input sequences also introduce higher computational demands and can significantly slow down training due to increased GPU memory consumption. 

In contrast to previous works~\cite{flamingo,openflamingo}, which employ a sequence length of 256, we opt for a sequence length of 128 to expedite the training process, within a limited computation budget. 
It is important to note that our model has the potential for further enhancement with the use of longer sequences if budget allows.

\begin{table*}[]
    \centering
    \footnotesize
    \scalebox{\GlobalTableRescale}
{
    \begin{tabular}{y{57}x{17}|lllll|lll|c}
    \toprule
    \bf Method  & \bf Shots & \multicolumn{5}{c}{\bf Captioning (CIDER)}  & \multicolumn{3}{c}{\bf Open-ended VQA} & \bf Average  \\
    & &TVC & MSVD & MSRVTT & YouCook2 & VATEX & MSRVTT & MSVD  & TGIF \\
    \midrule
    \multirow{3}{*}{\bf Flamingo-3B~\cite{flamingo}}&0 &-&-&-&55.8&40.1&11.0(-) & 27.5(-) &- & N/A\\
    &4&-&-&-&64.6&50.0&14.9(-) & 33.0(-)&- & N/A\\
    &32&-&-&-&76.7&59.2&25.6(-) & 42.6(-)&- & N/A \\
    \hline
    \bf \multirow{3}{*}{\ModelNameTwoB} & 0 & 6.0  &61.4  & 31.4 &17.9&33.5& 1.0(18.5) & 4.8(32.5) & 20.5(41.3) & 22.0\\
    &4& 6.7 &64.2& 34.2 &21.5&37.8&2.3(21.4) & 8.7(34.2)&23.2(45.2) & 24.8 \\
    &32 & 8.2 & 72.0 &37.5 & 31.5 & 36.8& 8.2(23.3)&11.5(33.4)& 28.2(46.6) & 32.3\\
    \hline
     \bf \multirow{3}{*}{\parbox[c]{3cm}{\ModelNameTwoB+\VideoDatasetName}} &
    0 & 6.7 &80.2&33.1&19.2&35.5&5.1(20.3)&15.2(34.3) & 21.5(43.3) & 27.1\\
    &4&8.3&82.7 & 42.3&28.4&39.9&8.4(22.5)&18.7 (35.2)&24.3(45.8)&31.6\\
    &32&12.2& 85.3 & 53.5 & 33.5 & 42.3 & 9.1(24.3)& 20.3(35.5)&30.4(49.3) & 35.8\\
    \bottomrule
    \end{tabular}
    }
    \vspace{-1em}
    \caption{
    \textbf{Video-text Task Comparison}.
    We assess these datasets using open-ended generation.
    Parentheses indicate evaluation via text similarity matching using Language Models.
    Except for YouCook2, we only utilize 3 frames per video.
    }
    \vspace{-1em}
    \label{tab:video_tasks}
\end{table*}

\subsection{Few-shot Evaluation on Vision-Language Tasks}
\label{sec:video_text_tasks}
\paragraph{Results on Image-Text Tasks.} Table~\ref{tab:sota} presents a comparative analysis with related works, demonstrating our state-of-the-art performance.
Our \ModelName{} outperforms Open-Flamingo~\cite{openflamingo} across nearly all benchmarks, and this is achieved with a substantially reduced sample size (130M vs. 180M) and fewer parameters. 
When using the same RedPajama-3B~\cite{redpajama} model as the language model, our model also outperforms Open-Flamingo substantially.
Furthermore, the incorporation of our proposed \VideoDatasetName~ dataset leads to even better results, underscoring the strength of the high-quality data.
It is worth noting that, during pre-training, our model is trained with only three frames, yet it delivers strong results even when evaluated with 32 shots.

\section{Training Details}

\paragraph{Results on Video-Text Tasks.}
We test our model on video captioning and video question-answering tasks.
When using an additional video dataset, we consistently observe enhanced model performance across various aspects.

However, it is important to note a significant limitation in the current MSVD and MSRVTT datasets: their ground truth quality is compromised due to being generated via rule-based methods rather than human annotation. 
Consequently, there are instances where our model generates more detailed and accurate predictions, which may not align with the ground truth. 
For example, while the ground truth is "Carve", our model predicts "Carving sculpture".
To address this challenge, we propose an alternative approach for evaluating VQA performance. 
This involves evaluating text similarity using a pre-trained Language model from NLTK~\cite{nltk}. 
The results of our evaluations are presented in the brackets of Table~\ref{tab:video_tasks}. 
Our findings indicate that, across a majority of the evaluated downstream tasks, our model consistently delivers robust performance.

\subsection{Analysis on Interleaved Data}
\label{sec:interleave_sampling_explore}

\begin{figure}
    \centering
    \includegraphics[width=\linewidth]{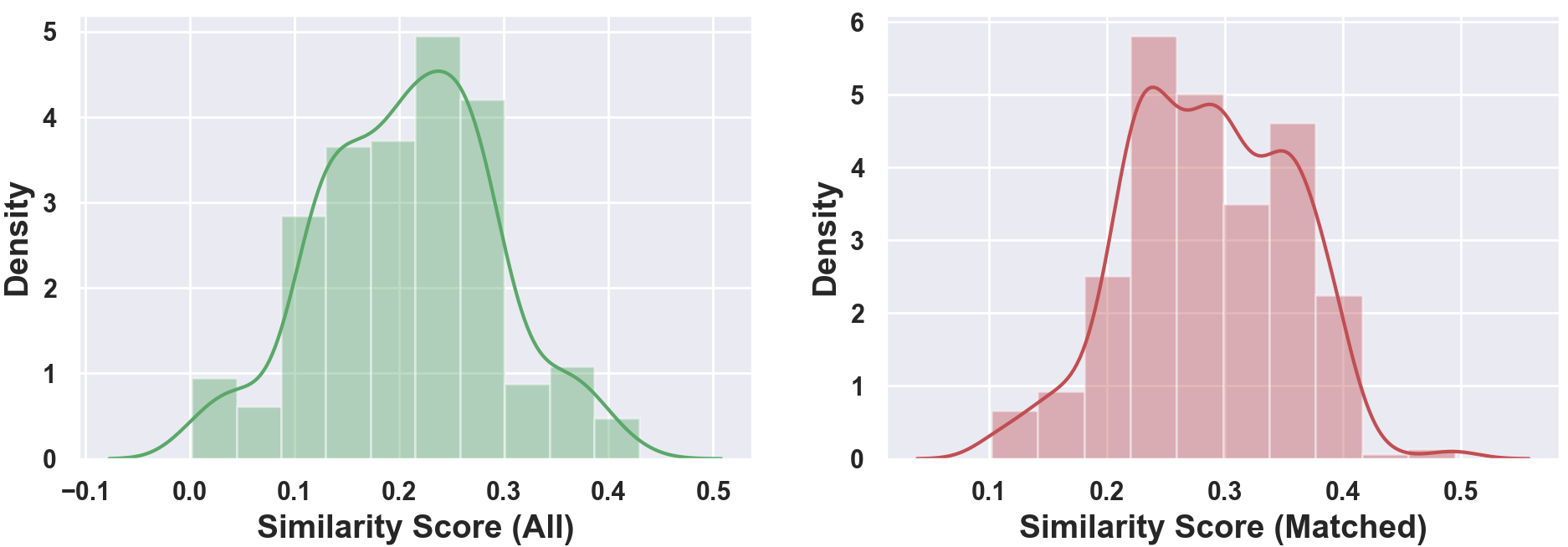}
    \caption{
    \textbf{The clip image-to-text similarity score distribution of MMC4.
    }
        Left is all pairs and the right is matched pairs.
    }
    \label{fig:mmc4similarity}
\end{figure}

\paragraph{Visualization of Interleaved Similarity Scores.}
We visualize the distribution of image-text similarity scores. The majority of similarity scores fall within the range of 0.2 to 0.4.
By varying the threshold from 0.18 to 0.30 while maintaining a dataset subset of 18 million, as consistent with our ablation setting, we present the results in Table~\ref{tab:thresh_abl}.

When the similarity threshold is below 0.22, frequent occurrences of gradient explosions hinder the successful training of our model. This issue predominantly arises from the disruptive influence of noisy data, significantly impairing training stability.
Furthermore, large thresholds (e.g., 0.28 and 0.30) yield suboptimal results on both COCO and FLICKR30K, partially due to a substantial dropout of interleaved samples. 
To address this, we experiment with replacing low-similarity captions with those generated from a pre-trained \ModelName, resulting in more stable training. 
We have noted that while replacing noisy captions contributes positively to performance enhancement, the correction of accurate pairs notably results in a significant decrease.

\paragraph{Analysis on Interleaved Data Sampling.}
Our empirical findings underscore that a larger number of shots significantly improves outcomes in the experiments conducted on interleaved frames. The predominant origin of most samples lies within raw images. To address potential bias and preserve the maximum number of frames, we implement a strategic approach: replacing captions with pre-trained data for images displaying exceptionally low similarity, accounting for approximately 5\% of the dataset. This procedural adjustment yields a marginal yet discernible enhancement in the model's overall performance.

\begin{table}[]
    \centering
    \footnotesize
    \begin{tabular}{c|ccccc}
    \toprule
      \bf Thresh  &\bf 0.18 &\bf 0.20 &\bf 0.24 &\bf 0.28 &\bf 0.30  \\
      \midrule
       COCO CIDER & N/A & N/A  &72.3 &64.7 &53.5 \\
     FLICKR CIDER & N/A &N/A &52.6&44.3&36.2\\
     \midrule
     \rowcolor{LightCyan} 
        COCO CIDER & 65.4 & 67.2  & 73.2 &63.3 &45.7 \\
        \rowcolor{LightCyan} 
     FLICKR CIDER & 42.3 & 48.4 &54.3&42.1&32.2\\
         \bottomrule
    \end{tabular}
    \caption{
    \textbf{The selection of similarity thresh score of interleaved affects the result.}
    N/A means the model fail into gradient explosion.
    The second line means we replace the noisy data with generated caption from pretrained \ModelName.
    }
    \label{tab:thresh_abl}
\end{table}

\begin{table}[]
    \footnotesize
    \centering
    \begin{tabular}{c|ccccc}
    \toprule
      \bf Subset   &\bf 0 &\bf 4 &\bf 8 &\bf 16 &\bf 32  \\
      \midrule
        Random Selection & 63.6 & 66.5 &71.3 &64.7 &60.5 \\
        Minimum Frames & 65.4&63.2&52.4&43.5&45.2\\
        Maximum Frames &64.1&66.8&73.5&75.5&75.0 \\
         \bottomrule
    \end{tabular}
    \caption{
    \textbf{Data quality highly affects the multiple-shots ability.}
    }
    \label{tab:subset_comparison}
\end{table}

\paragraph{Analysis of Interleaved Subsets:}
In this experiment, we closely examine three unique subsets derived from the MMC4 dataset:
\begin{itemize}
    \item A randomly sampled 4 million subset.
    \item 4 million samples with the highest count of images. 
    \item 4m samples that containing only a single frame.
\end{itemize}

To mitigate the potential issue of over-sampling documents with multiple frames, we limit the model training to a single epoch. 
The detailed results, as presented in Table~\ref{tab:subset_comparison}, reveal a significant performance gap, particularly favoring the subset enriched with the most frames. This notable disparity suggests that the effectiveness in few-shot learning largely stems from the integration of interleaved data combined with the strategic use of LLMs.

\subsection{Zero-shot Alignment Tasks}
In this experiment, we include zero-shot image classification and retrieval task.
We utilize the DataComp~\cite{datacomp} evaluation pipeline to test our model's capabilities. Specifically, the model's performance is evaluated across 38 datasets
without any training.
The result is shown in Table~\ref{tab:zero_shot_alignment}.

\begin{table}[]
    \centering
    \footnotesize
    \begin{tabular}{c|ccc}
        \toprule
        Method & VTAB & Retrieval & Average over 38 Datasets \\
        \midrule
        \ModelNameTwoB & 31.5 & 25.4 & 32.7  \\
        \ModelNameThreeB & 33.2 & 26.3  & 35.8 \\
        \bottomrule
    \end{tabular}
    \vspace{-1em}
    \caption{Zero-shot performance for alignment tasks in~\cite{datacomp}.}
    \vspace{-1em}
    \label{tab:zero_shot_alignment}
\end{table}

\section{Conclusion and Limitations}
In this work, we present a refined architecture aimed at incorporating contrastive loss into an existing autoregressive multi-modality model, \ModelName, tailored specifically for in-context learning across multiple modalities.
But in-context learning require high-quality interleaved data and there still no interleaved video-text dataset available.
To address this gap, we introduce \VideoDatasetName, a pioneering interleaved video-text dataset featuring comprehensive captions, marking a significant stride in this field's advancement.
Furthermore, the potential advantages of employing pre-training models in more downstream tasks, particularly in long text tasks, warrant further exploration. 
Our ongoing endeavor involves the forthcoming release of our trained models and the dataset, aiming to foster extensive research in this domain.

{
    \small
    \bibliographystyle{utils/ieeenat_fullname}
    \bibliography{main}
}

\clearpage

\appendix
\section*{Appendix} 

\section{Scaling up the Language Model}

\subsection{Data and Comparison}
Our endeavor in this experimental phase aimed at a significant upscaling of our language model, transitioning from the previously employed 2B and 3.4B configurations to a more expansive 8B setting.

To ensure a fair and consistent comparison across models, we augmented the data sample size within the DataComp subset from 130M to 180M, as outlined in the DataComp~\cite{datacomp} dataset. The associated data statistics, presented comprehensively in Table~\ref{tab:supp_8b_dataset_details}, showcase a substantial augmentation, specifically involving a 60M increase in the sample count within the Datacomp~\cite{datacomp} subset.

Our efforts weren't solely confined to data augmentation; we also incorporated a larger Mistral-7B language model~\cite{mistral} to achieve an overall parameter count of 8.1B. 
The comparison between our method and Open-flamingo, utilizing equivalent-scale pre-training data, consistently demonstrates the superior performance of our approach across diverse tasks.





\section{Training Methodology Details}

\subsection{Hyperparameters}

In this section, we delineate the crucial training specifics necessary for replication. The experimentation encompassed three variations in model size. Notably, larger models necessitated smaller batch sizes owing to GPU memory constraints. Our approach utilized deepspeed zero-stage 2 with fp16, while aligning gradient accumulation steps with the data type count. Detailed results in Table~\ref{tab:supple_hyperparameter_details}.

\subsection{Handling Exceptions}

Given the utilization of at least 8 nodes (64 GPUs) in our training setup, some challenges emerged during multi-node training. 
We encountered and addressed various issues:

\textbf{Skip Last Batch:} Random sampling occasionally resulted in unstable batches, leading to markedly high contrastive loss and language model instability.

\textbf{Loss Upper Bound:} To counter such occurrences, we implemented a moment value mechanism to track contrastive loss and language modeling loss. If the loss of batch surpassed the moment value by a certain threshold, we scaled down the loss, fostering more stable training.

\textbf{NAN Error Mitigation:} Periodically, the model encountered NAN errors without the possibility of recovery. To manage such instances, an automated job restart mechanism utilizing NCCL was employed, ensuring continuity in training despite potential failure points.

These strategies were pivotal in mitigating challenges inherent in multi-node training setups, ensuring a more stable and reliable training process despite the complexities introduced by distributed computing.

\begin{table}[]
    \centering
    \footnotesize
    \begin{tabular}{l|ll}
    \toprule
        Data Type & Dataset & Sample  \\
        \midrule
        \multirow{4}{*}{Image-Text} & CC3M~\cite{cc3m} & 0.75M \\
        &SBU~\cite{sbu} & 0.3M\\
        &LAION400M~\cite{laion400m} & 30M\\
        &DataComp1B~\cite{datacomp} & 116M\\
        \hline
        Interlevel Image-Text&MMC4~\cite{mmc4} &  30M \\
        \hline
        Video-Text & WebVid~\cite{webvid} & 2.5M \\
        \hline
        Interlevel Video-Text & \VideoDatasetName  & 0.9M \\
        \midrule
        Total & - & 180M \\
        \bottomrule
    \end{tabular}
    \vspace{-1em}
    \caption{
    \textbf{Statistics of the Pre-training Dataset
    for \ModelNameEightB}:
    Subsets are from the complete sets using clustering and filtering.
    }
    \label{tab:supp_8b_dataset_details}
\end{table}

\begin{table}[]
    \centering
    \footnotesize
    \begin{tabular}{c|cc}
    \toprule
       Method  & TGIF-MC & LSMDC-MC \\
       \midrule
       \ModelNameTwoB & 47.2 & 45.3\\ 
       \ModelNameThreeB & 50.3 & 46.6\\
       \ModelNameEightB & 53.4 & 50.2\\
       \bottomrule
    \end{tabular}
    \caption{Results on Video Multiple-Choice task on TGIF~\cite{tgif}  action split and LSMDC~\cite{lsmdc} val set with three frames.
    }
    \label{tab:supple_video_mc}
\end{table}

\begin{table*}[h]
    \centering
    \footnotesize
    \begin{tabular}{p{0.12\linewidth}|p{0.2\linewidth}|p{0.14\linewidth}p{0.14\linewidth}|p{0.14\linewidth}}
    \toprule
         &  & \texttt{\ModelNameTwoB} & \rule{0pt}{10pt}\texttt{\ModelNameThreeB} \rule[-5pt]{0pt}{0pt}
        &
        \rule{0pt}{10pt}\texttt{\ModelNameEightB} \rule[-5pt]{0pt}{0pt}
        \\ 
        \midrule
        \multirow{3}{*}{\rule{0pt}{10pt}\textbf{Model}} & \textit{Language Model Backbone} & OPT-IML-1.8B~\cite{optiml} & Redpajama-3B~\cite{redpajama} & Mistral-7B~\cite{mistral} \rule[-5pt]{0pt}{0pt}\\ \cline{2-5}
        \rule{0pt}{10pt}~ & \textit{Vision Model Backbone} & \texttt{openai/clip\allowbreak-vit-large\allowbreak-patch14} & \texttt{openai/clip\allowbreak-vit-large\allowbreak-patch14}   \rule[-5pt]{0pt}{0pt} & \texttt{laion/CLIP-ViT\allowbreak -H-14-laion2B\allowbreak -s32B-b79K}  \rule[-5pt]{0pt}{0pt}\\ \cline{2-5}
        \rule{0pt}{10pt}~ & \textit{Cross-Layer Interval} & 4 & 4 & 4\rule[-5pt]{0pt}{0pt}\\ \hline
        \multirow{5}{*}{\rule{0pt}{10pt}\textbf{Training}} & \textit{Sequence Length} & 128 & 128 & 128 \rule[-5pt]{0pt}{0pt}\\ \cline{2-4}
        \rule{0pt}{10pt}~ & \textit{Effective Batch Size} & 6144 & 3072 & 1536 \rule[-5pt]{0pt}{0pt}\\ \cline{2-5}
        \rule{0pt}{10pt}~ & \textit{Max Training Steps} & 200K & 200K &500K \rule[-5pt]{0pt}{0pt}\\ \cline{2-5}
        \rule{0pt}{10pt}~ & \textit{Weight Decay} & 0.1 & 0.1 & 0.1 \rule[-5pt]{0pt}{0pt}\\ \cline{2-5}
        \rule{0pt}{10pt}~ & \textit{Optimize}r & adamw(0.9, 0.999) & adamw(0.9, 0.999) & adamw(0.9, 0.999)\rule[-5pt]{0pt}{0pt}\\ \cline{2-5}
        \rule{0pt}{10pt}~ & \textit{Gradient Clipping} & 1.0 & 1.0 & 1.0\rule[-5pt]{0pt}{0pt}\\  \hline
        \multirow{5}{*}{\rule{0pt}{10pt}\textbf{Learning Rate}} & \textit{Initial Max} & 5e-5 & 5e-5 & 3e-5 \rule[-5pt]{0pt}{0pt}\\  \cline{2-5}
        \rule{0pt}{10pt}~ & \textit{Decay Schedule} & Cosine & Cosine &Cosine \rule[-5pt]{0pt}{0pt}\\ \cline{2-5}
        \rule{0pt}{10pt}~ & \textit{Linear warmup} Steps & 500 & 500 & 500 \rule[-5pt]{0pt}{0pt}\\ 
        \bottomrule
    \end{tabular}
\vspace{0.5em}
\caption{\textbf{The hyperparameters used in pre-training for three distinct \ModelName variations}.
The learning rate and batch size is smaller for \ModelNameEightB sine the GPU memory limitation is 32GB.
}
\label{tab:supple_hyperparameter_details}
\end{table*}

\begin{table*}[]
    \centering
    \footnotesize
    \scalebox{\GlobalTableRescale}
{
    \begin{tabular}{x{23}x{13}x{13}x{13}|x{13}x{13}|x{13}x{13}x{13}x{13}|c|cc|c}
    Method & Val Acc1 & Val Acc5 &Shots & \multicolumn{2}{c}{Captioning (CIDER)}  & \multicolumn{4}{c}{VQA} & \multicolumn{1}{c}{Classification} & \multicolumn{2}{c}{Retrieval(t2v,v2t)} & Average  \\
    & &  &  & COCO & FLICKR & ok-vqa & textvqa & vizwiz & vqav2 & hatefulmemes & COCO & FLICKR & (wo Retrieval)\\
    \hline
    \bf \multirow{5}{*}{\parbox[c]{2cm}{112,\\682.52M\\/2.32G,\\4d2h54m}} & \multirow{5}{*}{62.229} &\multirow{5}{*}{78.74} 
      & 0 & 60.5 & 41.1 & 14.6 & 17.9 & 9.1 & 22.3 & 48.3 & 22.8/21.2 & 50.4/40.5 & 29.3 \\
    &&& 4 & 67.3 & 49.5 & 21.6 & 16.2 & 20.5 & 42.0 & 48.1 & - & - &33.2\\
    &&& 8 & 72.3 &  55.3 & 20.8 & 16.0 & 24.7 & 42.6 & 48.6& - & - & 37.6\\
    &&& 16 & 67.8 & 49.3 & 15.5 & 12.4 & 28.7 & 32.2 & 52.5& - & - &32.3\\
    &&& 32 & 65.2 & 38.5 & 13.9 & 12.6 & 31.8 & 28.0 & 51.9& - & - &31.6\\
    \hline
    \bf \multirow{5}{*}{\parbox[c]{2cm}{121,\\682.52M\\
    /2.32G,\\4d1h13m}} &\multirow{5}{*}{62.329} & \multirow{5}{*}{78.2} 
    & 0 & 61.4 & 42.2 & 18.2 & 19.2 & 10.8 & 28.7 & 52.1 & 22.0/19.1 & 47.6/41.5 & 30.5 \\
    &&&4 & 68.5 & 51.6 & 20.8 & 16.4 & 23.5 & 40.6 & 50.6& - & - & \bf 36.3 \\
    &&&8 & 73.7 & 54.3 & 21.9 & 16.3 & 30.3 & 39.9 & 54.7& - & - & \bf 38.8\\
    &&&16 & 69.0 & 52.6 & 15.9 & 12.6 & 37.4 & 30.3 & 53.0 & - & - & \bf 36.3\\
    &&&32 & 66.4 & 47.3 & 13.8 & 13.2 & 39.8 & 26.3 & 49.5& - & - &\bf 34.2\\
    \hline
    \bf \multirow{5}{*}{\parbox[c]{2cm}{211,\\682.52M\\
    /2.32G,\\4d1h13m}} &\multirow{5}{*}{61.01} & \multirow{5}{*}{77.3} 
    & 0 & 66.4 & 45.7 & 17.9 & 19.2 & 9.5 & 36.2 & 49.0 & 22.2/20.8 & 48.8/43.8 & \bf 32.8 \\
    &&&4 & 71.0 & 52.4 & 21.5 & 16.3 & 22.5 & 42.0 & 50.3 & - & - & 34.1 \\
    &&&8 & 76.5 & 56.3 & 19.6 & 16.9 & 29.4 & 40.0 & 48.9& - & - & 38.2 \\
    &&&16 & 70.6 & 51.3 & 15.9 & 13.0 & 35.5 & 33.0 & 50.1 & - & - & 33.7 \\
    &&&32 & 65.7 & 39.4 & 12.9 & 14.6 & 40.0 & 29.1 & 50.7& - & - & 32.6 \\
    \hline
    \end{tabular}
    }
    \caption{
    \textbf{Ablation Study on Data Weights}:
    Interleaved data emerges as a more pivotal contributor compared to other data types.
    }
    \label{tab:supple_data_type_abl}
\end{table*}

\section{Video-based Multiple-choice Tasks}

In this experiment, we introduce an additional set of video tasks. Unlike open-ended video question answering, the multiple-choice task involves selecting an answer from a pool of candidates.

Precisely, our evaluation involves TGIF-MC~\cite{tgif} and LSMDC-MC~\cite{lsmdc} as benchmark datasets. To assess these multiple-choice tasks without any fine-tuning, we propose a similarity-based matching approach. Initially, we generate potential answers and then measure the maximum similarity against all candidates. The most similar candidate, identified using a Language Model from NLTK~\cite{nltk}, is selected. If the index matches the correct index, we mark it as a correct answer. Our findings indicate that our model demonstrates improved performance with larger model sizes in these tasks.

\section{Extended Results: Multi-shot Details}

In these experiments, we present \ModelName's performance across varying shot counts, ranging from 0 to 32 shots. For these ablation studies, our training utilized 64 interleaved tokens within an 18M subset.
For the Validation accuracy, we report the top1 and top5 accuracy.

\subsection{Data Weight Ablation}

This experiment delves into the data weight distribution among Image-text, Interleaved Image-text, and Video-Text. Prior to computing all losses with gradient accumulation, we applied corresponding data type weights. Table~\ref{tab:supple_data_type_abl} illustrates that \textbf{interleaved image-text pairs} notably influence downstream accuracy.
Driven by this observations, the data weight for interleaved data is 2 as default.

\subsection{PEFT method vs. Cross Attention}
Parameter-efficient fine-tuning methods like LORA~\cite{lora} are prevalent in LLM. The inquiry here is whether it is feasible to introduce LORA for model fine-tuning.

\begin{table*}[]
    \centering
    \footnotesize
    \scalebox{\GlobalTableRescale}
{
    \begin{tabular}{x{23}x{13}x{13}x{13}|x{13}x{13}|x{13}x{13}x{13}x{13}|c|cc|c}
    Method & Val Acc1 & Val Acc5 &Shots & \multicolumn{2}{c}{Captioning (CIDER)}  & \multicolumn{4}{c}{VQA} & \multicolumn{1}{c}{Classification} & \multicolumn{2}{c}{Retrieval(t2v,v2t)} & Average  \\
    & &  &  & COCO & FLICKR & ok-vqa & textvqa & vizwiz & vqav2 & hatefulmemes & COCO & FLICKR & (wo Retrieval)\\
    \hline
            \bf \multirow{5}{*}{\parbox[c]{2cm}{Cross-\\attention\\ Layer}} &\multirow{5}{*}{61.402} & \multirow{5}{*}{77.342}
    & 0&63.6 & 42.9& 10.9 & 14.9 & 6.5 & 11.5 & 50.3 & 22.2/20.1 & 49.8/40.1 &  28.6 \\
    &&& 4 & 66.5 &48.3 & 20.5 & 17.1 & 27.8 & 41.3 & 48.7 & - & - & \bf 38.6  \\
    &&& 8 & 71.3 &52.6 & 20.9 & 15.0 & 33.6 & 42.3 & 49.5& - & - & \bf 40.7 \\
    &&& 16 & 64.7&48.2 & 15.2 & 12.5&37.6 & 32.6 & 50.2& - & -  & \bf 37.3\\
    &&& 32 & 60.5&33.6 & 13.8 & 11.5& 40.3 & 26.9 & 49.7& - & - & \bf 33.7  \\
    \hline
    \bf \multirow{5}{*}{\parbox[c]{2cm}{LORA~\cite{lora}}} & \multirow{5}{*}{61.33} &\multirow{5}{*}{77.94} 
    & 0& 69.6 & 49.0 & 15.0 & 17.9 & 8.4 & 30.1 & 51.7 & 22.3/19.8 & 49.4/45.2 & \bf 31.5\\
    &&& 4 & 68.7 & 52.0 & 18.2 & 16.5 & 19.3 & 40.0 & 51.4 & - & - & 33.4\\
    &&& 8 & 73.3 & 54.7 & 19.0 & 15.9 & 24.6 & 39.9 & 49.7& - & - & 36.8 \\
    &&& 16 & 71.9 & 52.0 & 16.7 & 12.4 & 31.8 & 33.6 & 47.3& - & - & 33.4 \\
    &&& 32 & 72.9 & 43.5 & 13.4 & 12.5 & 34.1 & 31.2 & 48.8& - & - &  33.3 \\
    \hline
    \end{tabular}
    }
    \caption{
    \textbf{LORA Ablation Study}:
    The introduction of LORA does not significantly enhance the cross-attention within the interleaved framework.
    }
    \label{tab:supple_lora_ablation}
\end{table*}

\begin{table*}[]
    \centering
    \footnotesize
    \scalebox{\GlobalTableRescale}
{
    \begin{tabular}{x{23}x{13}x{13}x{13}|x{13}x{13}|x{13}x{13}x{13}x{13}|c|cc|c}
    Method & Val Acc1 & Val Acc5 &Shots & \multicolumn{2}{c}{Captioning (CIDER)}  & \multicolumn{4}{c}{VQA} & \multicolumn{1}{c}{Classification} & \multicolumn{2}{c}{Retrieval(t2v,v2t)} & Average  \\
    & &  &  & COCO & FLICKR & ok-vqa & textvqa & vizwiz & vqav2 & hatefulmemes & COCO & FLICKR & (wo Retrieval)\\
    \hline
    \bf \multirow{5}{*}{\parbox[c]{2cm}{3e-4\\682.52M\\
    /2.32G,\\3d7h12m}} &\multirow{5}{*}{61.402} & \multirow{5}{*}{77.342}
    & 0&63.6 & 42.9& 10.9 & 14.9 & 6.5 & 11.5 & 50.3 & 22.2/20.1 & 49.8/40.1 &  \bf 28.6 \\
    &&& 4 & 66.5 &48.3 & 20.5 & 17.1 & 27.8 & 41.3 & 48.7 & - & - & \bf 38.6  \\
    &&& 8 & 71.3 &52.6 & 20.9 & 15.0 & 33.6 & 42.3 & 49.5& - & - & \bf 40.7 \\
    &&& 16 & 64.7&48.2 & 15.2 & 12.5&37.6 & 32.6 & 50.2& - & -  & \bf 37.3\\
    &&& 32 & 60.5&33.6 & 13.8 & 11.5& 40.3 & 26.9 & 49.7& - & - & \bf 33.7  \\
    \hline
    \bf \multirow{5}{*}{\parbox[c]{2cm}{3e-5,\\682.52M\\
    /2.32G,\\3d7h12m}} &\multirow{5}{*}{55.78} & \multirow{5}{*}{72.96}
    & 0 & 37.8 & 21.6 & 21.8 & 12.3 & 8.4 & 42.2 & 48.0 & 1.6/1.8 & 2.7/4.1 & 24.0 \\
    &&&4 & 42.4 & 22.0 & 16.0 & 7.1 & 30.0 & 37.8 & 49.2 &  -&- & 25.9\\
    &&&8 & 25.0 & 16.7 & 14.7 & 5.5 & 35.8 & 32.4 & 49.1&  -&- & 21.9\\
    &&&16 & 12.4 & 9.9 & 1.9 & 3.0 & 40.8 & 24.8 & 51.5&  -&- & 15.5 \\
    &&&32 & 10.7 & 8.5 & 6.9 & 3.5 & 42.4 & 24.1 & 52.2&  -&-  & 16.0\\
    \hline
    \bf \multirow{5}{*}{\parbox[c]{2cm}{3e-6,\\682.52M\\/2.32G,\\3d12h29m}} & \multirow{5}{*}{49.70} &\multirow{5}{*}{69.70} 
    & 0 & 3.6 & 2.5 & 20.6 & 5.9 & 6.7 & 37.9 & 48.7&3.5/3.4&6.7/8.3 & 12.9 \\
    &&&4 & 4.6 & 3.5 & 14.2 & 4.4&30.0 & 34.7 & 45.3&  -&- & 15.2 \\
    &&&8 & 6.7 & 5.3 & 14.3 & 4.8&35.7&34.9 & 47.5&  -&- & 17.1 \\
    &&&16 & 8.0 & 6.4 & 13.0&4.7&40.8&32.7&49.1&  -&- & 17.8 \\
    &&&32 & 8.0 & 5.7 & 5.3&3.3&42.4&27.8&50.0&  -&- & 15.4 \\
    \hline
    \end{tabular}
    }
    \caption{
    \textbf{Learning rate ablation}.}
    \label{tab:supple_learning_rate}
\end{table*}

To this end, we introduce the LORA into text decoder.
Specifically, we keep the original cross attention but bring LORA additionaly.
We show the result in Table~\ref{tab:supple_lora_ablation}.
We find the PEFT method bring negative gains for cross-attention layers based method which already introduce a large number of trainable parameters.

\subsection{Learning Rate Ablation}
This experiment involved varying the learning rate from 3e-4 to 3e-6, and the reported results in Table~\ref{tab:supple_learning_rate} reveal a crucial finding. Extremely small learning rates significantly impair downstream task performance. For instance, a drop from 40.7 to 21.9 accuracy on 8 shots was observed.

As default, we use 5e-4 for \ModelNameTwoB and \ModelNameThreeB.
For \ModelNameEightB, we use learning rate 3e-4.

\subsection{Learning Rate Schedule}
Experimenting with four distinct learning rate schedules—Cosine, Constant, Cosine-w-restart, and Inverse Sqrt—unveiled substantial variations in final results, as demonstrated in Table~\ref{tab:supple_learning_rate_schedule}. Generally, the Cosine schedule yields the best results across most cases, thus adopted as the default scheduler due to its consistent performance.

\begin{table*}[]
    \centering
    \footnotesize
    \scalebox{\GlobalTableRescale}
{
    \begin{tabular}{x{23}x{13}x{13}x{13}|x{13}x{13}|x{13}x{13}x{13}x{13}|c|cc|c}
    Method & Val Acc1 & Val Acc5 &Shots & \multicolumn{2}{c}{Captioning (CIDER)}  & \multicolumn{4}{c}{VQA} & \multicolumn{1}{c}{Classification} & \multicolumn{2}{c}{Retrieval(t2v,v2t)} & Average  \\
    & &  &  & COCO & FLICKR & ok-vqa & textvqa & vizwiz & vqav2 & hatefulmemes & COCO & FLICKR & (wo Retrieval)\\
    \hline
    \bf \multirow{5}{*}{\parbox[c]{2cm}{Cosine\\682.52M\\
    /2.32G,\\3d7h12m}} &\multirow{5}{*}{61.402} & \multirow{5}{*}{77.342}
    & 0&63.6 & 42.9& 10.9 & 14.9 & 6.5 & 11.5 & 50.3 & 22.2/20.1 & 49.8/40.1 &  28.6 \\
    &&& 4 & 66.5 &48.3 & 20.5 & 17.1 & 27.8 & 41.3 & 48.7 & - & - & 38.6  \\
    &&& 8 & 71.3 &52.6 & 20.9 & 15.0 & 33.6 & 42.3 & 49.5& - & - & 40.7 \\
    &&& 16 & 64.7&48.2 & 15.2 & 12.5&37.6 & 32.6 & 50.2& - & -  & 37.3\\
    &&& 32 & 60.5&33.6 & 13.8 & 11.5& 40.3 & 26.9 & 49.7& - & - & 33.7  \\
    \hline
    \bf \multirow{5}{*}{\parbox[c]{2cm}{Constant(1e-4),\\682.52M\\
    /2.32G,\\2d21h40m}} &\multirow{5}{*}{57.00} & \multirow{5}{*}{73.95}
    & 0 & 35.5 & 30.3 & 22.2 & 14.6 & 9.3 & 42.9 & 46.4 & 3.4/3.8 & 8.2/9.0 & 26.0\\
    &&&4 & 52.3 & 52.3 & 19.9 &10.0 & 23.8 & 41.2 & 50.2&-&-&33.3\\
    &&&8 & 46.4 & 46.4 & 17.6&7.8 & 30.2 & 38.7 & 47.8&-&-&31.1\\
    &&&16 & 38.6 & 38.6 & 9.1 &5.7 & 32.6 & 35.2 & 52.9&-&-&26.6\\
    &&&32 & 24.2 & 24.2 & 6.3 & 4.2 & 35.3 & 24.8 & 48.6&-&-&19.8\\
    \hline
    \bf \multirow{5}{*}{\parbox[c]{2cm}{Cosine-w-restart,\\682.52M\\/2.32G,\\3d12h29m}} & \multirow{5}{*}{57.31} &\multirow{5}{*}{74.577} 
    & 0 & 36.1 & 34.3 & 19.0   &15.2&6.9 & 30.3 & 52.0&12.9/11.8&34.0/26.5&23.6\\
    &&&4 & 58.1 & 42.1 & 19.0&14.7&20.2 & 42.7 & 51.0&-&-&32.8\\
    &&&8  &  51.7& 38.9 & 11.8&10.7 & 26.8 & 34.2 & 52.8&-&-&29.0\\
    &&&16 & 50.1 & 28.1&5.1&7.2 & 30.4 & 34.1 & 49.9&-&-&25.8\\
    &&&32 & 21.5 & 4.5&2.7&3.7 & 36.2 & 25.3 & 50.5&-&-&15.7\\
    \hline
    \bf \multirow{5}{*}{\parbox[c]{2cm}{INVERSE SQRT,\\682.52M\\/2.32G,\\3d12h29m}} & \multirow{5}{*}{54.26} &\multirow{5}{*}{73.17} 
    & 0 & 28.3 & 15.7 & 14.6 & 6.6 & 4.8 &25.5& 50.9 & 1.8/2.2&4.4/4.7 & 15.9\\
    &&&4 & 24.8 & 11.0 & 8.3 & 3.7 & 25.5 & 30.8 & 54.5 &-&-& 17.3 \\
    &&&8 & 15.7 & 11.0 & 5.1 & 2.7 & 28.4 & 26.0 & 48.5 &-&-& 14.8\\
    &&&16 & 12.0 & 10.6 & 1.7 & 1.6 & 29.1 & 24.8 & 50.8&-&-&13.5\\
    &&&32 & 12.6 & 8.6 & 1.0 & 1.7 & 23.9 &23.3 & 46.4&-&-&11.9\\
    \hline
    \end{tabular}
    }
    \caption{
    \textbf{Learning rate schedule ablation}.
    }
    \label{tab:supple_learning_rate_schedule}
\end{table*}

\begin{table*}[]
    \centering
    \footnotesize
    \scalebox{\GlobalTableRescale}
{
    \begin{tabular}{x{23}x{13}x{13}x{13}|x{13}x{13}|x{13}x{13}x{13}x{13}|c|cc|c}
    Method & Val Acc1 & Val Acc5 &Shots & \multicolumn{2}{c}{Captioning (CIDER)}  & \multicolumn{4}{c}{VQA} & \multicolumn{1}{c}{Classification} & \multicolumn{2}{c}{Retrieval(t2v,v2t)} & Average  \\
    & &  &  & COCO & FLICKR & ok-vqa & textvqa & vizwiz & vqav2 & hatefulmemes & COCO & FLICKR & (wo Retrieval)\\
    \hline
    \bf \multirow{5}{*}{\parbox[c]{2cm}{No Memory,\\682.52M\\/2.32G,\\2d13h16m}} & \multirow{5}{*}{60.85} &\multirow{5}{*}{78.04} 
    & 0 & 65.6 & 46.2 & 15.8 & 18.3 & 7.9 & 34.0 & 47.7 & 23.0/20.0 & 51.5/43.3 &\bf 33.5\\
    &&&4 & 69.9 & 53.7 & 16.0 & 14.1 & 11.6 & 39.0 & 48.9& - & - &\bf 36.3\\
    &&&8 & 74.4 & 56.2 & 18.5 & 15.3 & 16.4 & 42.0 & 44.6& - & - &\bf 38.4\\
    &&&16 & 70.9 & 53.6 & 14.4 & 10.3 & 20.6 & 31.6 & 48.6& - & - &35.8\\
    &&&32 & 70.0 & 45.5 & 11.9 & 12.1 & 27.6 & 29.2 & 47.9& - & - & 34.9\\
    \hline
    \bf \multirow{5}{*}{\parbox[c]{2cm}{Retrieval Memory,\\682.52M\\
    /2.32G,\\3d7h12m}} &\multirow{5}{*}{63.012} & \multirow{5}{*}{79.319}
    & 0  &66.3&45.8&17.0&16.3&12.2&25.3&49.3&26.4/24.3&55.5/46.4&33.2\\
    &&&4 &68.3&54.2&15.3&13.2&14.4&35.0&40.8&-&- & 35.8 \\
    &&&8&72.3&55.6&19.3&13.3&17.8&40.3&49.3&-&-&36.5\\
    &&&16&73.4 & 52.4 & 10.3 & 13.4 & 18.2 & 35.3 & 50.3& - & - &\bf 36.2\\
    &&&32 & 70.7& 47.3 & 9.9 & 14.3 & 30.6 & 31.2 & 49.8& - & - &\bf 36.2\\
    \hline
    \end{tabular}
    }
    \caption{\textbf{Memory bank ablation}.}
    \label{tab:supple_memory_bank_abl}
\end{table*}







\section{Others}

\subsection{Data Selection Strategy for Image-Text}

The CC3M~\cite{cc3m}, SBU~\cite{sbu}, LAION400M~\cite{laion400m}, and DataComp1B~\cite{datacomp} datasets are widely accessible and commonly used as pre-training data. 
We include these datasets to facilitate easy replication. However, our observations suggest that while these image-text datasets are valuable, their impact on downstream performance is not as significant as that of interlevel image-text datasets.

To illustrate, utilizing the entire 130M data solely from DataComp~\cite{datacomp} resulted in minimal changes in performance. Furthermore, unlike previous methods like GiT~\cite{git} and BLip2~\cite{blip2}, we opted to exclude COCO and Visual Genome datasets during pre-training due to potential overlaps with downstream data. 
This strategic exclusion was implemented to minimize potential redundancy and maximize the distinctiveness of the learned representations.

\subsection{Exploring Longer Sequences with Memory Bank}

Managing long text input data holds significant importance in NLP. A recent approach, the Memorizing Transformer~\cite{memorizing_transformer}, introduces a memory bank to store segment tokens from the entire document. This technique enables the model to handle lengthy texts, surpassing even 10K tokens.

Following the principles of the Memorizing Transformer, our objective was to augment the interleaved data length. The results, detailed in Table~\ref{tab:supple_memory_bank_abl}, yielded intriguing observations:

\emph{i.} \textbf{Improved Validation Results:} Notably, the validation accuracy exhibited significant enhancement, e.g., from 63.012 to 60.85.
\emph{ii.} \textbf{Increased Computational Cost:} Introducing longer sequences incurred substantially higher computational demands.
\emph{iii.} \textbf{Limited Downstream Performance Gain:} Surprisingly, the performance gains on downstream tasks were relatively restricted and occasionally even negative.

Consequently, based on these findings, the Memorizing Transformer approach was not included in our final model configuration. However, this doesn't preclude the consideration of alternative methods to bolster the model's ability to process longer texts. There's potential to explore different approaches aimed at enhancing the model's handling of extended sequences.

\section{Training Progression}

In this section, we explore the training curves for three scales of \ModelName. Our findings reveal that larger Language Large Models (LLMs) show a reduced Language Model (LM) loss, particularly in processing interlevel image-text data, which underscores their proficiency in word prediction tasks.

However, a notable aspect was the slower convergence in contrastive loss for \ModelName compared to larger models like \ModelNameTwoB and \ModelNameThreeB, attributed to its smaller batch size. Additionally, hardware constraints were evident during training \ModelNameEightB on 32GB Tesla V100 GPUs, where the maximum batch size of 1536 impacted the convergence rate of contrastive loss.

These observations highlight the intricate balance between model size, batch size, and hardware limitations in the efficient training of LLMs.

\begin{figure*}[h]
    \centering
\includegraphics[width=\linewidth]{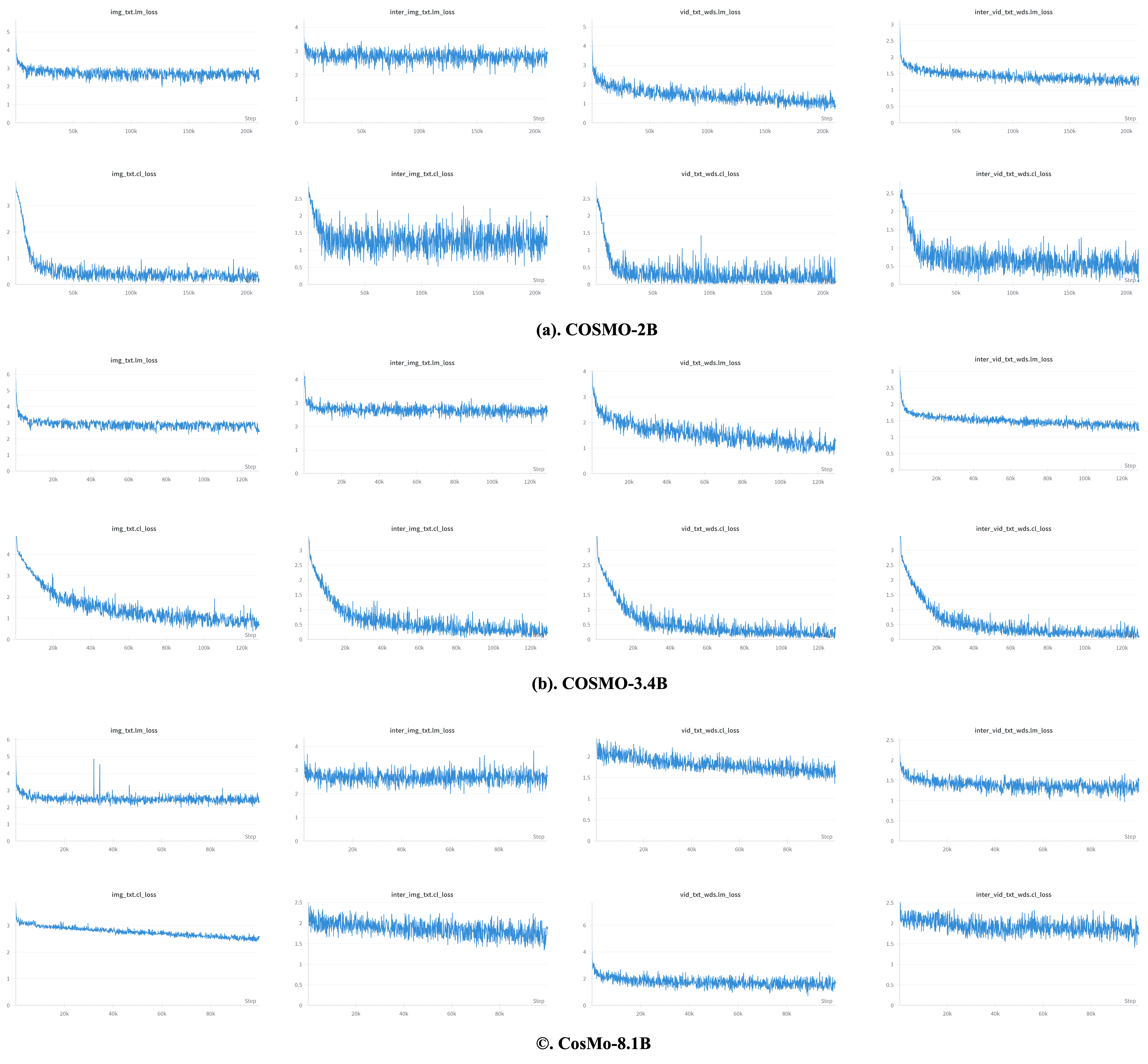}
    \caption{
    \textbf{Generally, Larger Language Models exhibit lower Language Modeling (LM) loss}.
    However, the convergence of contrastive loss tends to be slower owing to the smaller batch size.
    }
    \label{fig:supp_training_curve}
\end{figure*}


\end{document}